\documentclass[conference]{IEEEtran}
\IEEEoverridecommandlockouts
\usepackage{cite}
\usepackage{amsmath,amssymb,amsfonts}
\usepackage{algorithmic}
\usepackage{graphicx}
\usepackage{textcomp}
\usepackage{xcolor}

\usepackage{algorithm}

\def\BibTeX{{\rm B\kern-.05em{\sc i\kern-.025em b}\kern-.08em
    T\kern-.1667em\lower.7ex\hbox{E}\kern-.125emX}}
\begin{document}

\title{Privacy Preserving Stochastic Channel-Based Federated Learning with Neural Network Pruning\\
}

\author{\IEEEauthorblockN{1\textsuperscript{st} Rulin Shao}
\IEEEauthorblockA{\textit{Xi'an Jiaotong University} \\
Xi'an, China \\
shaorulin@stu.xjtu.edu.cn}
\and
\IEEEauthorblockN{2\textsuperscript{rd} Given Name Surname}
\IEEEauthorblockA{\textit{dept. name of organization (of Aff.)} \\
\textit{name of organization (of Aff.)}\\
City, Country \\
email address}
\and
\IEEEauthorblockN{3\textsuperscript{nd} Hui Liu}
\IEEEauthorblockA{\textit{Department of mathematics} \\
\textit{Mianyang Vocational College}\\
Mianyang, China \\
kaiyuanmifen@gmail.com}
\and
\IEEEauthorblockN{4\textsuperscript{th} Dianbo Liu}
\IEEEauthorblockA{\textit{CSAIL} \\
\textit{MIT}\\
Camrbidge, MA,USA \\
dianbo@mit.edu}

}

\maketitle

\begin{abstract}
Artificial neural network has achieved unprecedented success in a wide variety of domains such as classifying, predicting and recognizing objects. This success depends on the availability of big data since the training process requires massive and representative data sets. However, data collection is often prevented by privacy concerns and people want to take control over their sensitive information during both training and using processes. To address this problem, we propose a privacy-preserving method for the distributed system, Stochastic Channel-Based Federated Learning (SCBF), which enables the participants to train a high-performance model cooperatively without sharing their inputs. We design, implement and evaluate a channel-based update algorithm for the central server in a distributed system, which selects the channels with regard to the most active features in a training loop and uploads them as learned information from local datasets. A pruning process is applied to the algorithm based on the validation set, which serves as a model accelerator. In the experiment, our model presents equal performances and higher saturating speed than the Federated Averaging method which reveals all the parameters of local models to the server when updating. We also demonstrate that the converging rates could be increased by introducing a pruning process. 
\end{abstract}


\section{Introduction}
In this section, we introduce the background of the related works with problems emerged in these domains, and briefly demonstrate our proposed model in solving these problems. 

\subsection{Federated Learning}
In conventional deep learning, all training data are shared to the central server who performs the analysis, and the clients who contribute the data have no control over it. That means, each client may have to upload their sensitive data to the server and do not know what the data is actually used for. Furthermore, the learned model is generally not directly available to the client so that they have to reveal the inputs to the cloud when using the model \ref{b32}, risking privacy leakage in both training and using processes. Federated learning can address this problem by introducing some algorithmic techniques that distribute its learning process to local devices so that the clients could keep their data private and obtain a local model for future use. 

Federated optimization has been studied by J. Konečný et al. \cite{b16,b19} for distributed optimization in machine learning. This work introduces a setting for distributed optimization where none of the typical assumptions \cite{b19} are satisfied, making federated learning a practical alternative to current methods. The proposed framework is different from conventional distributed machine learning \cite{b15, b21, b22, b23, b24, b25} for the large number of clients, highly unbalanced and non-i.i.d. data available on each client, and relatively poor network connections \cite{b20}. In the problem of the last constraint, Konečný, Jakub, et al. \cite{b20} have proposed two approaches, structured updates and sketched updates, to reduce the uplink communication costs. McMahan et al. \cite{b14, b30} advocates the Federated SGD (Stochastic Gradient Descent) and Federated Averaging algorithms as practical methods for the federated learning of deep networks based on iterative model averaging. Opposed to protect a single data point’s contribution in learning a model \cite{b18}, R. C. Geyer \cite{b17}  proposed an algorithm for client sided federated optimization in order to hide clients’ contributions during training. Further, methods to strengthen the reliability of federated learning, such as Secure Aggregation \cite{b27}, essentially require some notion of synchronization on a fixed set of devices, so that the server side of the learning algorithm only consumes a simple aggregate of the updates from many users \cite{b26}. Applications based on federated learning algorithm have been proposed in several domains, ranging from content suggestions \cite{b29} to next word prediction \cite{b28}. Research done by Bagdasaryan et al. \cite{b31} focus on the vulnerability of federated learning. This work shows that federated learning algorithm is vulnerable to a model-poisoning attack, which is different from poisoning attacks that target only the training data.  

Besides the direct leakage of privacy as mentioned before, participants in the distributed system may indirectly reveal some information about the sensitive data via the weights uploaded to the server in the training process. Addressing the direct as well as the indirect privacy leakage, we developed the Stochastic Channel-Based Federated Learning (SCBF) method which enables the local participants manipulate their data confidentially while benefit their model's performance from the server with only small proportion of the local trained gradients stochastically revealed to the central model.

\subsection{Differential Privacy Preserving} 

\begin{figure*}[!t]
\centering
\includegraphics[width=6in]{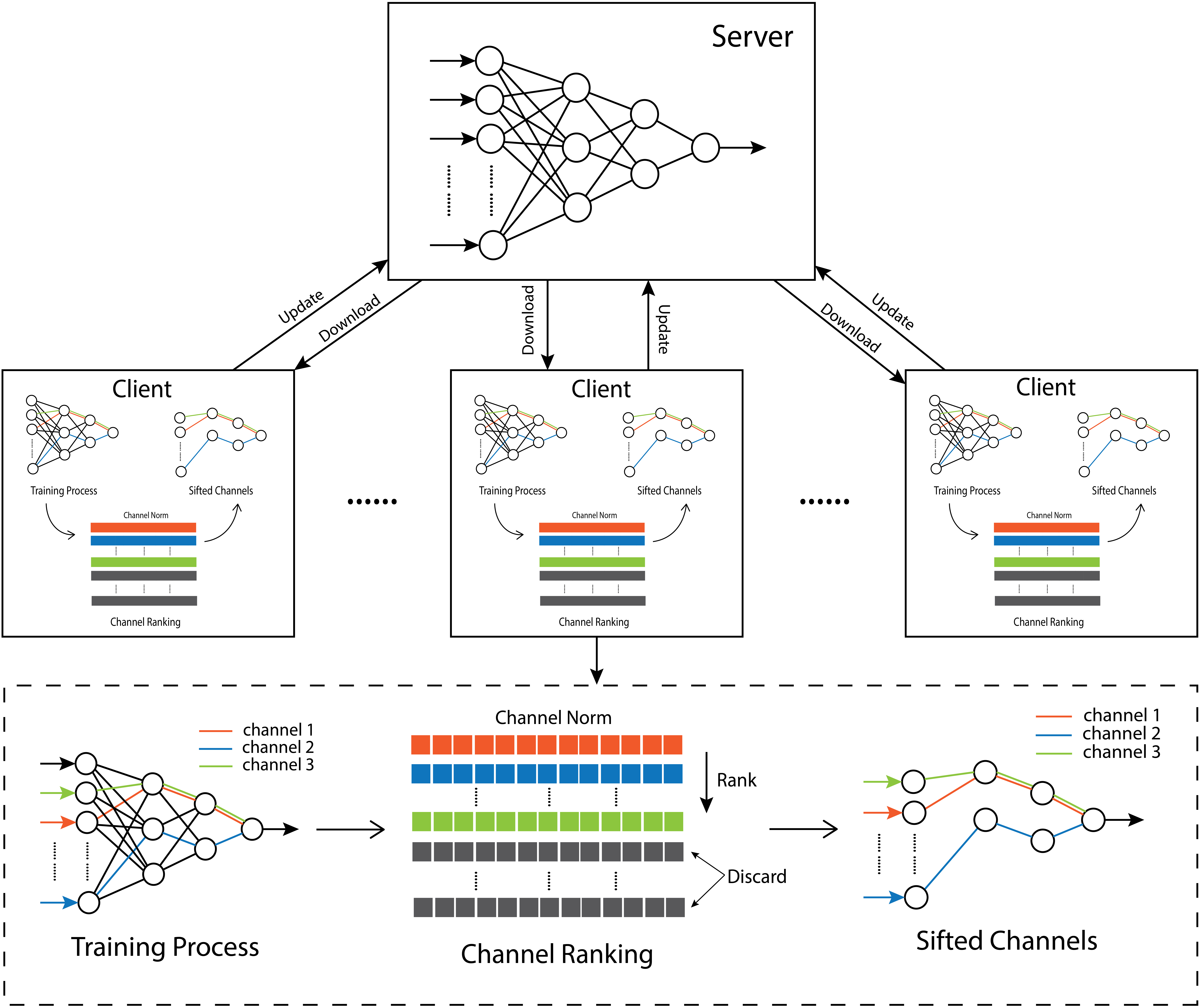}
\caption{SCBF Model}
\label{Fig: Stochastic Channel-Based Federated Learning with pruning}
\end{figure*}

Differential privacy \cite{b6, b7, b10, b12, b13} as a strong criteria for privacy-preserving is defined that a model is differential private if the probability of a given output doesn't primarily depend on the involvement of a data point in the inputs \cite{b32}. It is useful because conventional deep learning has great privacy concerns which may prevent the company from collecting data for model training. A model-inversion attack may extract parts of the training data through a deep learning network, as Fredrikson et al. \cite{b1} demonstrated. One might attempt to reduce the risk of privacy leakage is adding noise to the parameters that result from the training process. However, it's hard to achieve a balance between performances and privacy preserving since stronger noise brings protection for privacy as well as worse performances. Therefore, we seek ways that help to preserve local privacy during the training process. 

Addressing the problems mentioned above, the Stochastic Channel-Based Federated Learning (SCBF) realizes the function of differential privacy preserving by protecting the two sources of potential privacy leakage from federated learning: the actual values of uploaded gradients from the local participants and the mechanism these gradients are chosen \ref{b32}. By setting a threshold to select the parameters of gradients channel-wise, the actual values uploaded to the server are stored in a sparse tensor that processed from Stochastic Gradient Descent (SGD), a stochastic training process which has already been used for many privacy preserving cases \cite{b8, b9}. Besides, the participant could independently choose the update rate for their models, thus making it hard to track the selection of the channels that used for update, especially when they are trained individually using different datasets through stochastic ways.

\subsection{Reducing the Model Size}
Training a model with privacy-preserving methods could be time-consuming, especially when training sets are enormous. Addressing this problem, we introduce a neural network pruning process to SCBF that could prune off the redundant nodes in the neural network based on the validation set, thus saving a lot of time. This work is done circularly in the first several global loops until the distributed system reaches a suitable scale, enabling SCBF with Pruning (SCBFwP) to efficiently learn from the datasets.

Neural network pruning is not a brand-new concept. Yang \cite{b34} proposed a way to prune connections based on weights magnitude. Y. He \cite{b35} used a channel pruning method to accelerate deep convolutional neural network. H. Han and J. Qiao \cite{b36} introduced a growing-and-pruning approach for FNN (fuzzy neural network). In the work done by Srinivas \cite{b37}, a systematic way was proposed to prune one neuron at a time, addressing the problem of pruning parameters in a trained NN model.

Our experiment shows that neural network pruning could efficiently speed up the training process as well as the saturating of performance. And better performance is achieved when tuning the pruning proportion to cut off the redundant neurons in several training loops.

%The pruning process in SCBFwP also serves a privacy-preserving function because it increases the difficulty for attackers to track the local information by reducing the model scale according to validation sets.

\subsection{Contribution of Our Work}
The Stochastic Channel-Based Federated Learning (SCBF) could address both direct and indirect privacy leakage concerns, which trains client models on the local datasets and improve the server performance by uploading a proportion of gradients. The clients do not need to share their datasets with the server model during training process nor reveal their inputs when using it. Moreover, the inverse-model attack which analyzes information from the uploaded parameters could be obstructed by the stochastic nature of our upload algorithm taking advantage of Stochastic Gradient Descent (SGD). Our model achieves better performance than Federated Averaging, a state-of-the-art federated learning method, without uploading model weights. And the SCBF with Pruning (SCBFwP) could speed up the saturating of performance and save executing time. Better performance could be achieved by tuning the pruning proportion to cut off redundant neurons in the network.

\section{Material and Methods}

In this section, we introduce the experiment setup and the material used in our trail. The details of Stochastic Channel-Based Federated Learning (SCBF) algorithm is also demonstrated in this section with a specific focus on the server update procedure, which involves uploading process that rises privacy concerns. Besides, a concise explanation for the pruning mechanism is added in this section.

\subsection{Distributed Learning Setting}
We propose a privacy-preserving federated learning method based on the neural network, which could be executed on a distributed system, for example, mobile devices, to achieve collaborative deep learning goals with little risk of privacy leaks. Each device trains its model on the local dataset for several epochs in each global loop, and only stochastically upload a small percent of the model weights to the server in order to achieve good performance in the server without sharing the local data nor the overall model weights.

In our trial, we implement a distributed system with 5 clients contributing to one server. Pre-experiments are conducted to decide the proper structure for the proposed model. Through manual tuning, we find out that the model achieves best performance with high efficiency using 3 layers. So for each local client, we construct an artificial neural network for binary prediction of mortality with 3 fully connected layers and 64, 32, 1 neurons in corresponding layers using ReLU activation at hidden layers and sigmoid activation at the output layer. Besides, we add a dropout layer between the second and third hidden layers for reducing overfitting.

Regarding parameters of communication between server and clients, the download rate is set to 100 percent for each client model, supposing the parameters of the server are shared publicly. The update rate is set to 30 percent for both channel-based federated learning method and distributed selective SGD method. To enhance the influence of the latest update parameters, we choose 0.8 as the decay rate. As for the training process, we train each model for 100 global loops and 5 epochs in each loop with batch size set to 32. 

We use the stochastic gradient descent (SGD) algorithm to optimize our neural networks. Concerning the configuration of SGD, the learning rate is a hyperparameter that controls how much to adjust the model in response to the estimated error each time the model weights are updated. Our experiments on testing models with various learning rates suggest that the proper learning rate is around 0.01 to guarantee both good performance and stable results. 

In addition to the configuration of the model, the importance of performance measurement has long been recognized by academics. When it comes to the assessment of a classification model, we can count on the AUCROC and AUCPR. Higher the AUC, better the model is at distinguishing between patients with the demise and survival.

\subsection{Stochastic Channel-Based Federated Learning}

Based on the observation that different features do not contribute equally to the training process and the importance of each feature may vary from one dataset to another, Stochastic Channel-Based Federated Learning (SCBF) is a privacy-preserving approach which seizes the most vital information from the local training results only by uploading a small fraction of gradients stochastically. The intuition behind this method is that the biological neural circuit follows the Law of Use and Disuse and the strongest neurons for an individual is those constitute an active circuit in its learning process, suggesting that the neurons in one artificial neural network is not independent through a specific training process. Thus we could consider the collaborative effect of neurons in each channel (similar to the biological neural circuit) when selecting parameters for server update: if a channel of neurons change a lot in a training loop, we can assume it be a strong neural circuit in the network, corresponding to a sensitive feature in the input sets; While the neural channels with little change in one training loop should be regarded as deteriorated ones, whose information could be kept private with little effect on the server's final performance. Choosing the channels with the most substantial variation enables SCBF to only upload a small percent of the gradients in each training loop while achieving comparable accuracy to the Federated Averaging (FA) method without uploading the integrate local weights to the server, as will be demonstrated in the result part.

The update algorithm plays an essential role in SCBF. In each global loop, SCBF computes the norms of channels in gradients that result from the local training process, calculates the $\alpha$-percentile of the norms and then sifts out the channels with greater variation in the gradients than the percentile,  where $\alpha$ is the update rate set by the local participant. The sifted parameters are  used for the server update.

To facilitate the description of the algorithm, suppose there are $N$ features as input and a $L$-layer deep neural network is conducted with $m_1,m_2,\cdots,m_L$ neurons in each layer. For convenience sake, denote $m_0=N$ as the input dimension. Denote the wight matrix as $W=[W_1,W_2, \cdots, W_L]$ and bias matrix as $B=[B_1,B_2, \cdots, B_L]$. The shapes of weight matrix and bias matrix could be expressed as follows:
$$W_l=(w_{ij}^l)_{m_{l-1}\times m_l}$$
$$B_l=(b_i^l)_{m_l}$$
where $ l=1,2,\cdots,L$ and $m_l$ stands for the number of neurons in $l$-th layer. 

The update algorithm includes five steps: 
\subsubsection{Train Local Model}
The local models are trained separately on its own datasets and each model results a gradient matrix showing the change in weight matrix during each training loop. Denote the gradient matrix as $G$ and it shares the same shape with weight matrix $W$. Since the influence from the bias matrix is negligible compared to the weight matrix, the changes in bias is omitted for the efficience sake.

\subsubsection{Compute Channel Norms}
Considering that a channel must go through a neuron in each layer and correlate to a $L$-dimensional vector comprising the  index of these neurons, the results of channels' norm could be saved in a $L$-dimensional tensor $T$, each element of which equals a channel norm. The shape of $T$ should be:
$$T=(t_{i_1i_2\cdots i_L})_{i_1i_2\cdots i_L=1}^{m_1m_2\cdots m_L}.$$
Denote $c^{(i)}=[g_0^{(i)},g_1^{(i)},\cdots,g_L^{(i)}]$ as $i$-$th$ channel where $\vec{i}=[i_1,i_2,\cdots,i_L]$ is the index of tensor which correlates the neurons this channel goes through in each layer; The Euclidean norm of each channel is calculated by
$$n^{(i)}=\| c^{(i)} \|_2  =\sum_{j=0}^L(g_j^{(i)})^2,$$ 
and  is saved in the $L$-dimensional tensor $T$:
$$T_{i_1,i_2,\cdots,i_L}= n^{(i)}=\| c^{(i)} \|_2  =\sum_{j=0}^L(g_j^{(i)})^2.$$

\subsubsection{Sort Norms}
Given a fixed upload rate $\alpha$ (also called update rate in this paper), we could straighten the gradient tensor to a vector and sort it, computing the $\alpha$-quantile $q_{\alpha}$ as a threshold for the channel selection.

\subsubsection{Process Gradients}
There are two ways to process the gradients: 

\begin{itemize}
\item{Negative Selection:} Discard the channels whose norms are below the $\alpha$-quantile and select the rest parameters for update.
\item{Positive Selection:} Select the channels whose norm is above $q_{\alpha}$ with the rest parameters set to zeros.
\end{itemize}

In our trail, both selection methods work well. On the consideration that different neural channels may include same neurons, the positive selection tends to behave better than the negative selection by a preference to upload more parameters with the same update rate. Take the positive selection for example, for each element $T_{t_1,t_2,\cdots,t_L}$ in tensor $T$ which corresponds to a specific channel, process the gradients as regard to the rank of this channel's norm, as shown in the following form:

\[
(G_i)_{t_i} = 
\begin{cases}
(G_i)_{t_i}&\text{if } T_{t_1t_2\cdots t_L}>q_{\alpha},\\
0&\text{else}.
\end{cases}
\]

\subsubsection{Update Server}
In the end, upload the processed gradient matrix $\tilde{G}$ to the server and the server updates by adding gradients $\tilde{G}$ to its original weights.

\begin{algorithm}
\label{Alg: update}
\caption{Pseudocode of server update}
\begin{algorithmic}
\REQUIRE Training set $\bf{(X,y)}$, update rate $\bf{\alpha}$, local model, server model

\STATE Train the local model on $\bf{(X,y)}$ and save the gradients $\bf{\Delta W}$;
\STATE Calculate the Euclidean norm of each channel and save the results in tensor $\bf{T}$;
\STATE Straighten tensor $\bf{T}$ to a vector and compute its $\bf{\alpha}$-quantile $\bf{t_{\alpha}}$;
\STATE Select the channels from gradients $\bf{\Delta W}$ according to $\bf{t_{\alpha}}$\ and get processed gradients $\tilde{\bf{\Delta W}}$;
\STATE Update non-zero part of $\tilde{\bf{\Delta W}}$ to the server's weights.

\RETURN Updated server model
\end{algorithmic}
\end{algorithm}

The server update algorithm is executed every global loop, and our experiment shows that even uploading a 10\% percent of local channels could the server get comparative performances to the Federated Averaging methods with higher speed to reach saturation. And before the next training loop begins, the local model download the server's latest weights. The download rate is set to 100\% since we suppose that the server weights could be shared publicly, which could be adjusted according to the application scenarios.

\begin{figure}[!t]
\centering
\includegraphics[width=3.50in]{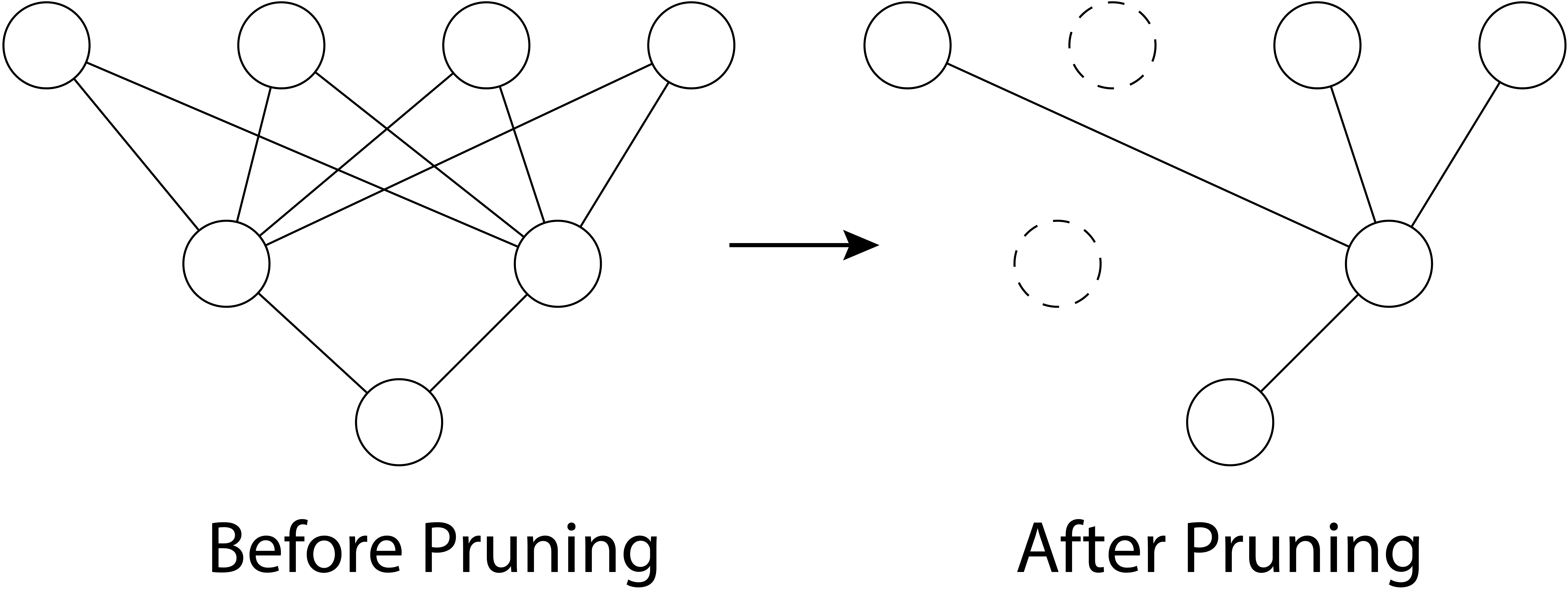}
\caption{Neural Network Pruning}
\end{figure}

\subsection{Pruning Process}

Models with relatively complex structures tend to be more suitable for solving complicated problems, while simple-structured models may suffer from underfitting and some other problems [38]. However, time-consuming and overfitting are two main problems a complex model may lead to. Addressing these problems, we introduced a pruning process to cut off the redundant neurons in the network, saving much of the time during the training process. 

Given the fact that each neural network has a computation process consisting of multiplication, addition and activation, neurons with mostly zeros output may have little effect on the output of subsequent layers, not to mention the final results \ref{b33}. Removing these redundant nodes from the model will do little harm to the accuracy of the network but save abundant executing time.

Average Percentage of Zeros (APoZ) \cite{b33}, which measures the percentage of zeros appeared in the activations of a neuron under the ReLU mapping, is used to evaluate the redundancy of neurons in the network. Denotes $CO_c^{(i)}$ as the output of $c$-$th$ neuron in $i$-$th$ layer. Let $M$ denotes the output dimension, and $N$ denotes the total quantity of validation examples. $APoZ_c^{(i)}$ of the $c$-$th$ neuron in $i$-$th$ layer is defined as:
$$APoZ_c^{(i)}=APoZ(O_c^{(i)})=\frac{\sum_k^N\sum_j^Mf(O_{c,j}^{(i)}(k)=0)}{N\times M}$$
where $f(\cdot)=1$ if true, and $f(\cdot)=0$ if false.

Stochastic Channel-Based Federated Learning with Pruning (SCBFwP) decides which neurons to be pruned according to APoZ using validation sets: those  having the highest APoZ will be pruned, the number of which is a fixed percentage of the total number of neurons left in each global loop.

\begin{algorithm}
\label{Alg: overall}
\caption{Pseudocode of SCBFwP}\label{update}
\begin{algorithmic}
\REQUIRE Models of local clients, model of the central server, update rate $\bf{\alpha}$, pruning rate $\bf{\theta}$, total pruned fraction $\bf{\theta_{total}}$, number of global loops, clients number $\bf{K}$

\FOR{global loops}

\FOR{each client}
\STATE Train the client model on local datasets;
\STATE Select channels according to the update rate and process the gradients $\bf{\Delta W_k}$;
\STATE Upload the processed gradients $\bf{\tilde{\Delta W_k}}$ to the server;
\ENDFOR

\STATE Update the server weights $\bf{W}$ with processed gradients from each client: 
\STATE $\bf{W} \leftarrow \bf{W} + \sum_{k=1}^{\bf{K}}{\bf{\tilde{\Delta W_k}}}$;

\IF{pruned fraction $\leq$ total pruned fraction}
\STATE Prune $\bf{\theta}$ of the server model according to validation set;
\STATE Prune each local model according to the structure of pruned server;
\ENDIF

\ENDFOR
\RETURN A distributed system with learned models
\end{algorithmic}
\end{algorithm}

\subsection{Dataset for Experiment}
Data used in our experiment was provided by hospitals, comprising 30760 admissions with status information represented by alive or expired. To explore the relationship between mortality and admissions, we develop a model that takes the medications as inputs and predictions of binary mortality as output. The cohort is managed in 2917 different medicines in total. Information on whether a patient takes each of the medicines after admission are adopted as binary input features. We use 60$\%$ of the dataset for training, 10$\%$ as the validation set, and 30$\%$ as the test set. The training set is equally divided into five parts as local training sets.

\section{Results}

In our experiment, we implement a distributed system with 5 clients contributing to one server. Pre-experiments are conducted to decide the proper structure for the proposed model. Through manual tuning, we construct an artificial neural network for each client with 3 fully connected layers and 64, 32, 1 neurons in corresponding layers using ReLU activation at hidden layers and sigmoid activation at the output layer. Besides, we added a dropout layer between the second and third hidden layers.

Regarding parameters of communication between server and clients, the download rate was set to 100 percent for each client model and the update rate is set to 30 percent for stochastic channel-based federated learning method. We choose 0.8 as the decay rate to enhance the influence of the latest update parameters. Each model is trained for 100 global loops and 5 epochs in each loop with batch size set to 32. 

We use the stochastic gradient descent (SGD) algorithm to optimize our neural networks. Our experiments on testing models with various learning rates suggest that the proper learning rate is around 0.01 to guarantee both good performance and stable results. Besides, AUCROC and AUCPR are used to evaluate our model.

30760 admissions with status information from hospitals are used in our experiment, of which 60\% are used for training, 10\% are used for validating and 30\% are used for testing or evaluating. The cohort is managed in 2917 different medicines in total and whether a patient takes each of the medicines after admission are adopted as binary input features.

\subsection{Stochastic Channel-Based Federated Learning}
The Stochastic Channel-Based Federated Learning (SCBF) method computes the norms of channels in gradients result from the local training output after each global loop, calculates the $\alpha$-percentile of the channel norms and then sifts out the channels that have greater variation in the gradients than the percentile for the server update. In this method, the server seizes the information from those uploading channels with biggest variation, thus achieving comparative performances with the stat-of-the-art method which has to convey the entire local weights to the server when updating.

Fig \ref{Fig: SCBF_Model} shows the relationship between the server and clients and demonstrates the process of server update.

\begin{figure*}[!t]
\centering
\includegraphics[width=7in]{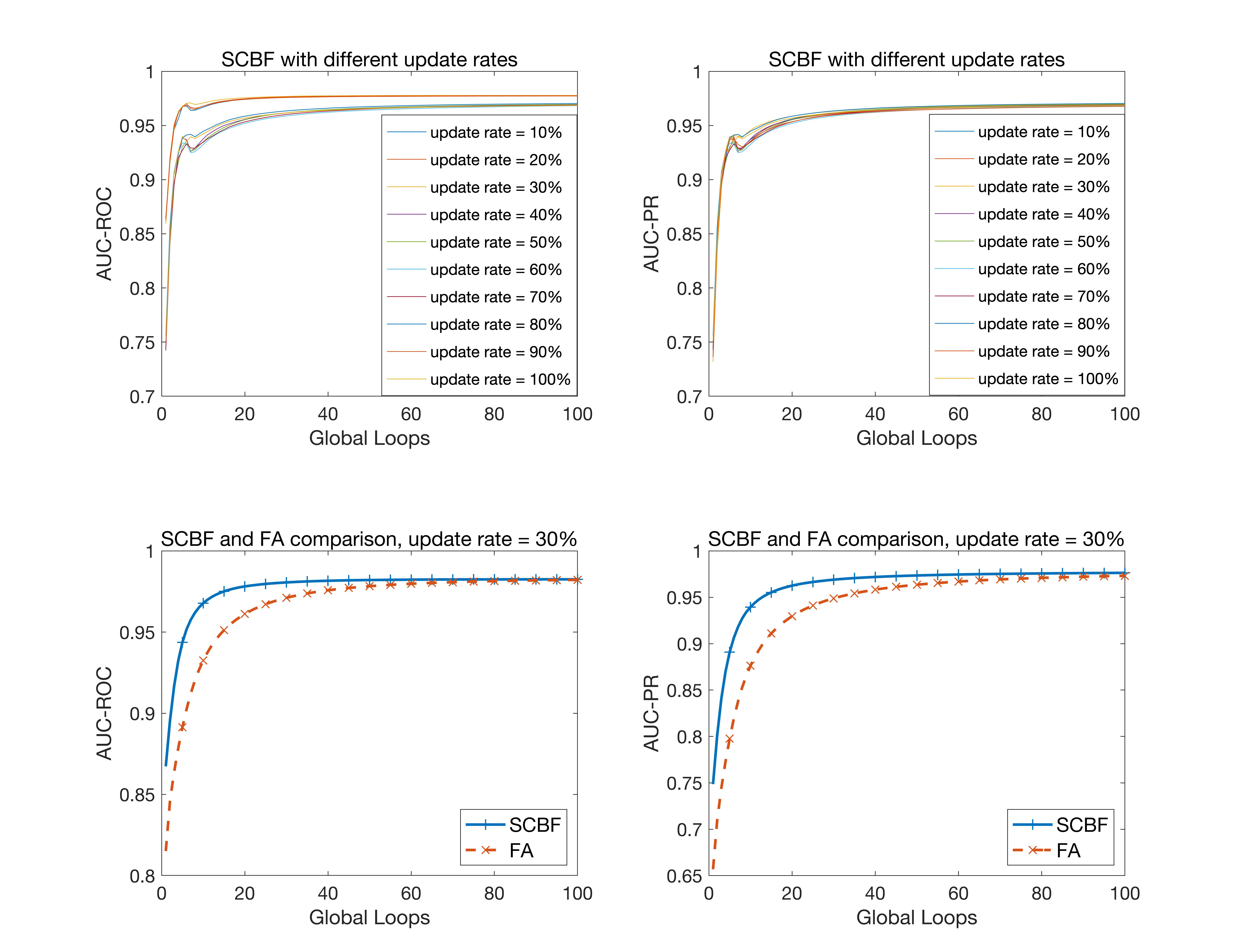}
\caption{Performances of SCBF. The first two graphs show performances of SCBFwP with different update rates and the comparison between the performances of SCBF and Federated Averaging are revealed in the last two graphs. The left column  shows the AUCROC performances and the right column shows AUCPR performances.}
\label{Fig: alpha and model comparison}
\end{figure*}

\begin{table}[htbp]
\caption{Saturated performances of SCBFwP with different update rates}
\begin{center}
\begin{tabular}{ccc}
\hline
\textbf{Update Rate}&\textbf{AUCROC} &\textbf{AUCPR}\\
\hline
10\% & 0.9776 & 0.9695 \\
20\% & 0.9772 & 0.9686 \\
30\% & 0.9777 & 0.9697 \\
40\% & 0.9768 & 0.9604 \\
50\% & 0.9780 & 0.9695 \\
60\% & 0.9774 & 0.9682 \\
70\% & 0.9774 & 0.9688 \\
80\% & 0.9781 & 0.9703 \\
90\% & 0.9774 & 0.9676 \\
100\% & 0.9775 & 0.9685 \\
\hline
\end{tabular}
\label{Tab: update rate}
\end{center}
\end{table}

\subsection{Performance of SCBF}

The update rate controls how many channels are selected whose non-zero part is uploaded to the server in each global loop, playing a vital role in affecting the final performance. To choose a suitable update rate for our distributed system, we implement SCBF models with different update rates ranging from 10\% to 100\%. The neural network pruning is used in this step to accelerate the training process. The performances are plotted in the first row of Fig \ref{Fig: alpha and model comparison}. The result shows that even with 10\% channels uploaded to the server, the SCBF model achieves an AUCROC of 0.9776 and an AUCPR of 0.9695, and this result even outperforms the model when sharing all parameters with the server. It confirms the intuition behind SCBF: the importance of a feature differs when training on different datasets, and we could extract important information from the channels that features with biggest variations go through. We could infer that less than 10\% of the channels contain the most fundamental information and ignoring the rest information does little harm to the learning of models. Besides, using a wide range of upload rates only leads to a 0.01319 amplitude in AUCROC and a 0.02739 amplitude in AUCPR, which facilitates the configuration process with a stable high performance.

To show the effectiveness of SCBF method compared to Federated averaging (FA), which implements the federated learning by averaging the gradients obtained from local training processes and is widely used in distributed systems. We set the update rate as 30\% for SCBF and conduct both methods for federated learning on the same datasets for 100 global loops without pruning. As shown in Fig \ref{Fig: alpha and model comparison}, our model reaches saturation at the $20th$ global loop, much faster than the FA which saturates at the $60th$ global loop. The performance of SCBF keep exceeding that of FA. In the $4th$ global loop, SCBF achieves 0.05388 higher in AUCROC and 0.09695 higher in AUCPR than FA. After 100 global loops, the AUCROC of SCBF is 0.0033 higher than FA and the AUCPR of SCBF is 0.0032 higher than FA. We could conclude that our method achieves comparative performance to the Federated Averaging method with higher saturating speed. What stands out in our method is that our model doesn't have to reveal the overall local models to the server and makes it hard for attackers to track the channels we choose.

\begin{figure*}[!t]
\centering
\includegraphics[width=7in]{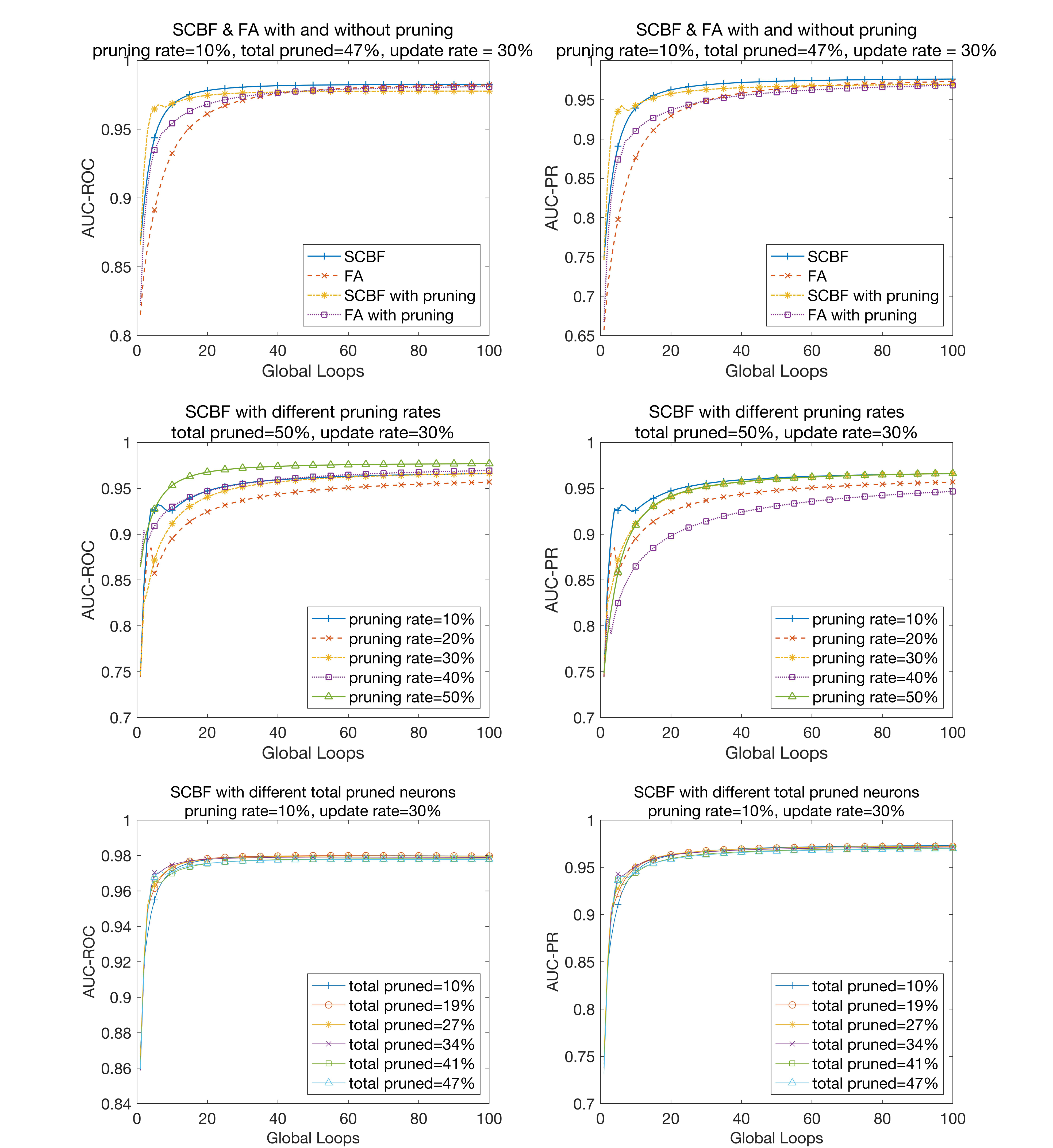}
\caption{Performance of SCBFwP. The first two graphs show the comparison between SCBF and FA with and without pruning. The third and fourth graphs show the performance of SCBFwP with different pruning rates. And performance of SCBFwP with different total pruned neurons are shown in the last two graphs. The left column reveals AUCROC performances and the right column reveals AUCPR performances.}
\label{Fig: pruning comparison}
\end{figure*}

\begin{table}[htbp]
\caption{Saturated performances of SCBF compared with FA}
\begin{center}
\begin{tabular}{ccc}
\hline
\textbf{Method}&\textbf{AUCROC} &\textbf{AUCPR}\\
\hline
SCBF & 0.9825 & 0.9763 \\
Federated Averaging & 0.9821 & 0.9731 \\
\hline
\end{tabular}
\label{Tab: comparison between SCBF and FA}
\end{center}
\end{table}

As shown in Fig \ref{Fig: download and upload}, when the upload rate for channels is set to 30\%, the parameters uploaded to the server is 45\% using positive selection. With half of the parameters unrevealed to the server, the model achieves better performance and higher saturating speed.

\subsection{Performance of SCBFwP}
To speed up the training process and reduce the size of the neural network, we conduct network pruning for several loops after pre-training the model. It is important to train a small-scaled deep learning model with high processing speed. In our trail, we set the pruning rate for each global loop to 10\%, which demonstrates the proportion of neurons to be pruned in the training loop. The total proportion of neurons to be pruned in the first several loops is set to 47\%, which decides the final scale of pruned model. Fig \ref{Fig: pruning comparison} compares the AUCROC and AUCPR values of both models, the SCBF and FA, with and without pruning. The results show that network pruning could speed up training process and accelerate convergence while maintaining higher performances. The results also show, as expected, that pruning 47\% neurons from the network will decrease the final performance due to the simplified model structure. The AUCROC for SCBF with pruning is reduced by 0.0048 and the AUCPR for it is reduced by 0.006814. We can observe a reduction of 0.0012 in AUCROC and 0.0047 in AUCPR for Federated Averaging method compared to FA with pruning. The reduction in performances is negligible in many application situations but the acceleration in both saturating speed and training speed is quite beneficial, the latter of which will be discussed in the following section.

\begin{table}[htbp]
\caption{Saturated performances of SCBF and FA with and without pruning}
\begin{center}
\begin{tabular}{ccc}
\hline
\textbf{Method}&\textbf{AUCROC} &\textbf{AUCPR}\\
\hline
FA & 0.9821 & 0.9731 \\
SCBF & 0.9825 & 0.9763 \\
FAwP & 0.9809 & 0.9683 \\
SCBFwP & 0.9776 & 0.9694 \\
\hline
\end{tabular}
\label{Tab: with and without pruning}
\end{center}
\end{table}

Moreover, the best performance is achieved by SCBF after 100 loops of training with 0.9825 for AUCROC and 0.9763 for AUCPR. The highest evaluation in the first 5 loops is obtained by the SCBF model with pruning. The results demonstrate that SCBF is a reliable choice for federated learning and SCBF with pruning method might be a better choice for whom preferring a quicker saturating speed.

In the first row of Fig \ref{Fig: pruning comparison}, there is an obvious decline in the performance for the SCBF with pruning, which indicates an over-pruned phenomenon for our trail. So there is a tradeoff between time efficiency and the final accuracy. And by tuning the pruning rate for each global loop and the total pruned rate of the model, we could achieve better performance because if only the redundant neurons are pruned, the model could promote its learning efficiency without remembering useless information. 

Also as regard to the stability of our model with the pruning rate and total pruned fraction (also called total pruned rate), we execute the models of SCBFwP controlling the variate. Firstly, we fix the total pruned fraction as 50\% and run the programs with different pruning rates ranging from 10\% to 50\%. As shown in the figure, with the pruning rate increasing, the final performance gets better and saturates quicker for most circumstances, but there are also exceptions regarding the high performances of 10\% pruning rate for both AUCROC and AUCPR, and lower performance of 40\% pruning rate for AUCPR. In the third row of Fig \ref{Fig: pruning comparison}, we fix the pruning rate to 10\% and execute pruning for different times ranging from 1 to 6. The total pruned fractions are calculated and annotated in the corresponding labels. The figure shows that SCBF gets better performance when reducing the times of pruning. The results with a fixed pruning rate is more stable than those with a fixed total pruned rate, indicating that people should pay more attention to the selection of pruning rate for each step when building models and it is stable for a SCBF model to adjust the times of neural network pruning. So after choosing a suitable pruning rate, we could appropriately increase the loops in which the model is pruned to shorten the executing time with little effect on the final performance.

\begin{table}[htbp]
\caption{Saturated performances of SCBFwP when total pruned proportion is fixed and pruning rate for each training loop changes}
\begin{center}
\begin{tabular}{ccc}
\hline
\textbf{Pruning Rate/Loop}&\textbf{AUCROC} &\textbf{AUCPR}\\
\hline
10\% & 0.9765 & 0.9661 \\
20\% & 0.9730 & 0.9568 \\
30\% & 0.9763 & 0.9662 \\
40\% & 0.9693 & 0.9465 \\
50\% & 0.9769 & 0.9663 \\
\hline
\end{tabular}
\label{Tab: pruning rate}
\end{center}
\end{table}

\begin{table}[htbp]
\caption{Saturated performances of SCBFwP when pruning rate for each training loop is fixed and total pruned proportion changes}
\begin{center}
\begin{tabular}{ccc}
\hline
\textbf{Total Pruned Proportion}&\textbf{AUCROC} &\textbf{AUCPR}\\
\hline
10\% & 0.9769 & 0.9731 \\
19\% & 0.9797 & 0.9722 \\
27\% & 0.9795 & 0.9725 \\
34\% & 0.9789 & 0.9714 \\
41\% & 0.9781 & 0.9703 \\
47\% & 0.9778 & 0.9697 \\
\hline
\end{tabular}
\label{Tab: total pruned fraction}
\end{center}
\end{table}

\begin{figure}[!t]
\centering
\includegraphics[width=3in]{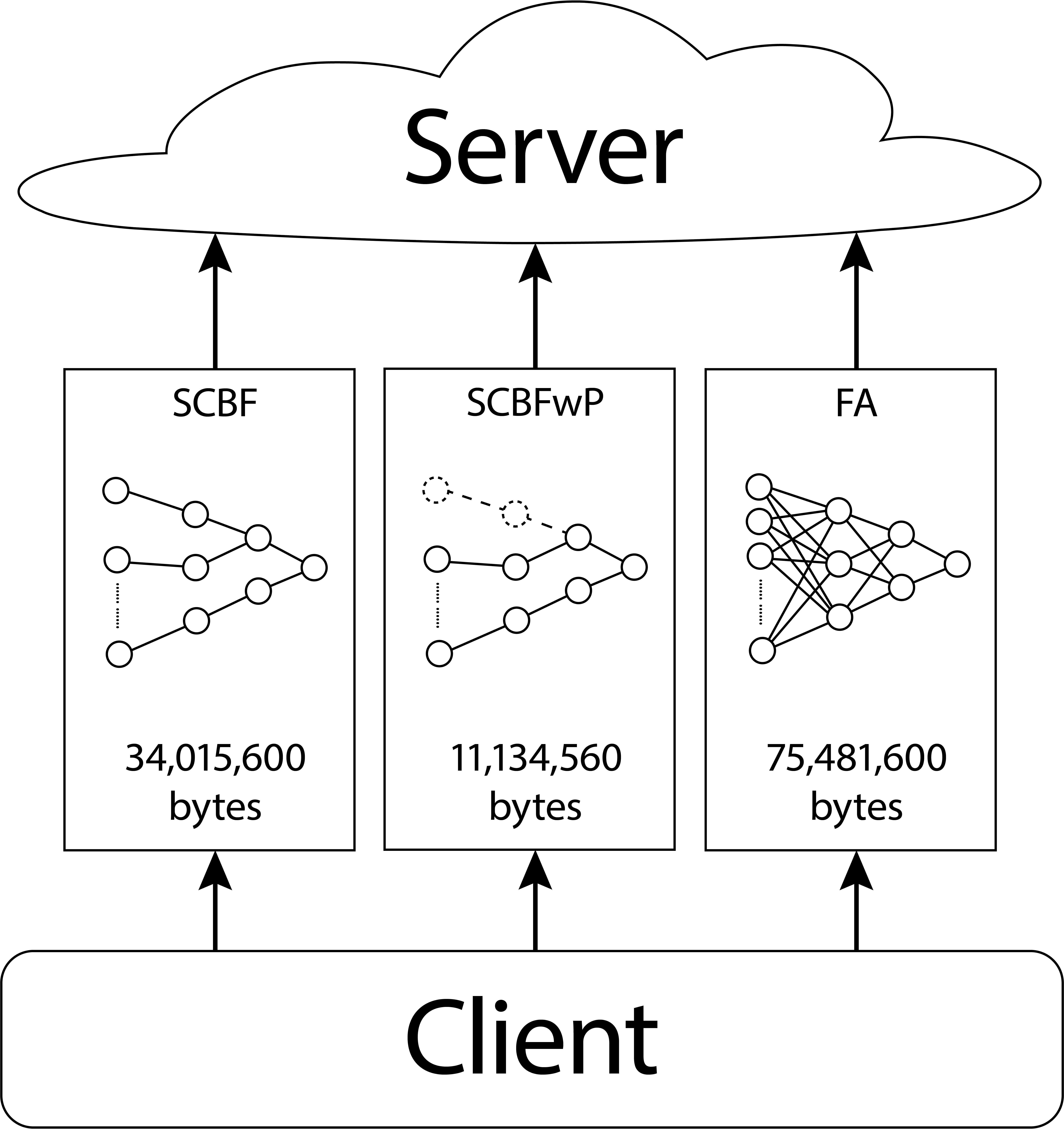}
\caption{Transinformation for upload processes using different methods. The SCBFwP could save 85\% of the transinformation compared to FA. And the SCBF could save 55\% by contrast with FA.}
\label{Fig: download and upload}
\end{figure}

As shown in Fig \ref{Fig: download and upload}, the SCBFwP could save 85\% of the transinformation compared to Federated Averaging. And for SCBF, when the upload rate for channels is set to 30\%, the parameters uploaded to the server is 45\% using positive selection. 

\subsection{Running Time}
SCBF preserves the privacy by adding a channel-based upload algorithm, which will lead to an increased burden of calculations when applied to a complex neural network, which, however, could be addressed by introducing pruning process in several global loops. To illustrate this, the time consumed by SCBF and FA before and after pruning described in the last section are listed in Table \ref{Tab: time consumed by methods}. As could be seen in the table, pruning process could reduce 57\% of the time for SCBF and 48\% of the time consumed by FA.

\begin{table}[htbp]
\caption{Time consumed by SCBF and FA with and without pruning}
\begin{center}
\begin{tabular}{cc}
\hline
\textbf{Method}&\textbf{Time (second)} \\
\hline
Federated Averaging & 8679 \\
Federated Averaging with Pruning & 4508 \\
SCBF & 19696 \\
SCBFwP & 8469 \\
\hline
\end{tabular}
\label{Tab: time consumed by methods}
\end{center}
\end{table}

\begin{table}[htbp]
\caption{Time consumed by SCBFwP with different update rates}
\begin{center}
\begin{tabular}{cc}
\hline
\textbf{Update Rate}&\textbf{Time (second)} \\
\hline
10\%&8339\\
20\%&8545\\
30\%&8469\\
40\%&7987\\
50\%&8359\\
60\%&12577\\
70\%&9278\\
80\%&11462\\
90\%&13169\\
100\%&13030\\
\hline
\end{tabular}
\label{Tab: time consumed with different alpha}
\end{center}
\end{table}

\begin{table}[htbp]
\caption{Time consumed by SCBFwP with different pruning rates for each loop or different total pruned proportions}
\begin{center}
\begin{tabular}{cc|cc}
\hline
\textbf{Pruning Rate/Loop}&\textbf{Time (second)}&\textbf{Total Pruned}& \textbf{Time (second)} \\
\hline
10\%&11144&10\%&25755\\
20\%&8561&19\%&22717\\
30\%&11852&27\%&17579\\
40\%&8389&34\%&15909\\
50\%&12000&41\%&8050\\
-&-&47\%&8469\\
\hline
\end{tabular}
\label{Tab: time consumed with different pruning rates}
\end{center}
\end{table}

Table \ref{Tab: time consumed with different alpha} shows that models with lower update rates tend to consume less time than those with larger update rates, indicating that choosing a lower rate for update could better preserve the privacy as well as save time.

As regard to the time consumed by models with different pruning rates and different total pruned rates, Table \ref{Tab: time consumed with different pruning rates} shows that different pruning rates for each global loop can equally save the time. And the model will consume more time if too few neurons are pruned due to the executing time of pruning process. With a fixed pruning rate, time consumed by the model tends to decrease by reducing the model size.

\section{Conclusion}
We proposed a privacy-preserving approach for distributed systems whose models are trained based on any type of neural network. Our methodology develops a channel-based update algorithm for the server, which enables the system to achieve a state-of-the-art performance without forcing the participants to reveal their inputs nor the local model weights to the server. Addressing the concern raised by inverse-model attacks, our model uploads a fraction of channels in the gradients from local models to the server and achieves better performance with 10\% channels uploaded than 100\% on the medical datasets, reducing the redundancy of gradients while preserving the privacy. Moreover, we introduced a neural pruning process to the model, which could accelerate the training process and saturating speed of performances with little sacrifice of the final performances.

Differential privacy could be further conducted on our models to evaluate the privacy-preserving ability quantitatively.

%\section*{Acknowledgment}

\end{document}